# COMPILED: Deep Metric Learning for Defect Classification of Threaded Pipe Connections using Multichannel Partially Observed Functional Data

Juan Du[a,b,c]*, Yukun Xie[a,b,c], Chen Zhang[d]

*[a] Smart Manufacturing Thrust, Systems Hub, The Hong Kong University of Science and Technology (Guangzhou), Guangzhou, China; [b] Academy of Interdisciplinary Studies, The Hong Kong University of Science and Technology, Hong Kong SAR, China; [c] The Hong Kong University of Science and Technology, Hong Kong SAR, China; [d] Department of Industrial Engineering, Tsinghua University, Beijing, China*

*Contact by Email: juandu@ust.hk

# COMPILED: Deep Metric Learning for Defect Classification of Threaded Pipe Connections using Multichannel Partially Observed Functional Data

### Abstract

In modern manufacturing, most products are conforming. Few products are nonconforming with different defect types. The identification of defect types can help further root cause diagnosis of production lines. With the sensing technology development, process variables evolved as time changes, which can be collected in high resolution as multichannel functional data. These functional data have rich information to characterize the process and help identify the defect types. Motivated by a real example from the threaded pipe connection process, we focus on defect classification where each sample is represented as partially observed multichannel functional data. However, the available samples for each defect type are limited and imbalanced. The functional data is partially observed since the pre-connection process before the threaded pipe connection process is unobserved as there is no sensor installed in the production line. Therefore, the defect classification based on imbalanced, multichannel, and partially observed functional data is very important but challenging. To deal with these challenges, we propose an innovative classification approach named as COMPILED based on deep metric learning. The framework leverages the power of deep metric learning to train on imbalanced datasets. A novel neural network structure is proposed to handle multichannel partially observed functional data. The results from a real-world case study demonstrate the superior accuracy of our framework when compared to existing benchmarks.

*Keywords:* imbalanced classification, partially observed functional data, functional neural network, deep metric learning, threaded pipe connection process.

## 1. Introduction

Threaded pipe connections have extensive applications in various industries, particularly in petroleum drilling and transportation. The importance of high-quality threaded pipe connections (*VAM book*, 2023) in ensuring safety during petroleum transportation is underscored by the fact that defective threaded pipe connections incur an annual cost of approximately half a billion USD (Guangjie *et al.*, 2006). Notably, a significant portion of the quality issues in threaded pipes arise from nonconforming connections. Thus, meticulous examination of the connection quality is a critical factor in determining the overall quality of threaded pipes.

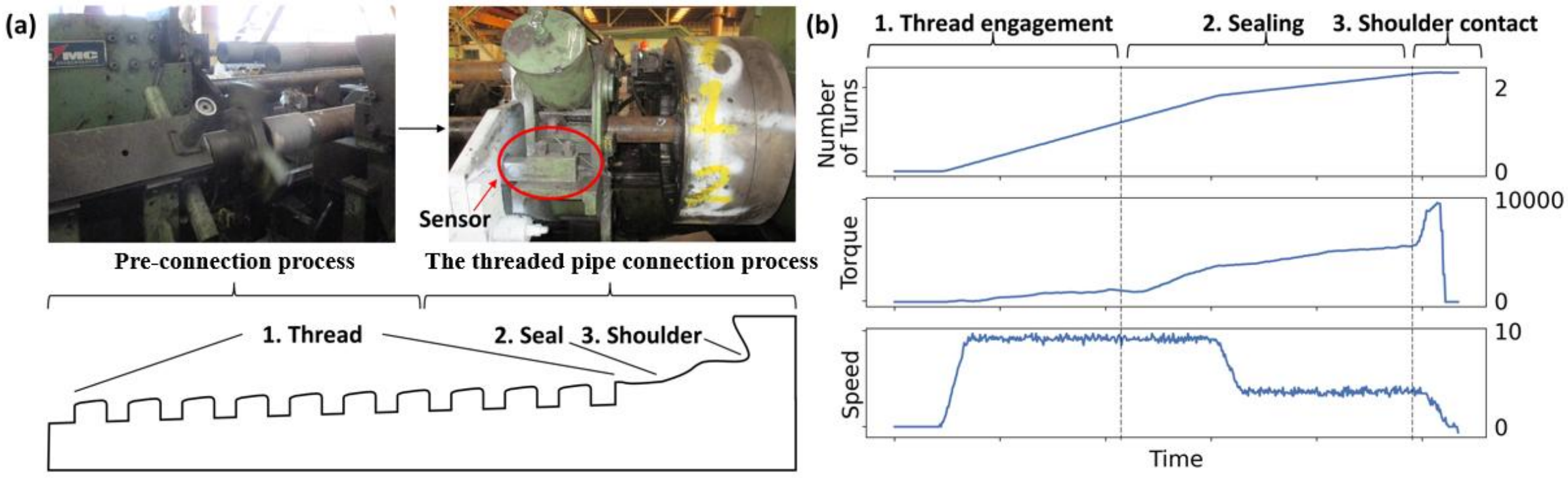

Figure 1. (a) Threaded pipe pre-connection process (left), connection process (right), and structure of threaded pipe connections (Du *et al.*, 2017); (b) Sensor signals collected in the threaded pipe connection process.

The threaded pipe connection process depicted in Figure 1 (a) encompasses a pre-connection process followed by a pipe connection process (Honglin *et al.*, 2014). During the pre-connection process, most threads are initially screwed, with the final connection achieved in the subsequent threaded pipe connection process. As depicted in Figure 1 (b), sensors are positioned on pipe connection machines to record process variables, including number of turns, torque, and connection speed. These sensor signals provide a wealth of process-related information and reveal distinct phases, such as thread engagement, sealing, and shoulder contact, within the threaded pipe connection process. Consequently, the connection quality inspection can be achieved according to these functional data from sensors. Currently, practitioners within the petroleum industry rely on functional data from sensors to manually identify nonconforming pipe connections (*VAM book*, 2023), and further expert intervention is needed to further identify the specific defects and implement corresponding remedial measures. Therefore, establishing a classification methodology based on the functional data from sensors is desired to automatically detect nonconforming connections and identify the connection defects, thereby enabling further root cause diagnosis of the manufacturing process.

To construct an automatic and accurate defect classification method, the imbalanced classification problem needs to be considered. In a regular production line, the samples for conforming and nonconforming products are highly imbalanced. There are a limited number of samples for each defect category, which poses significant challenges, such as overfitting, for the classification problem. Take the threaded pipe connection as an example, the ratio between conforming connections and each type of nonconforming connections reaches 20:1. The samples from a threaded pipe connection assembly line illustrate the imbalanced samples in Figure 2.

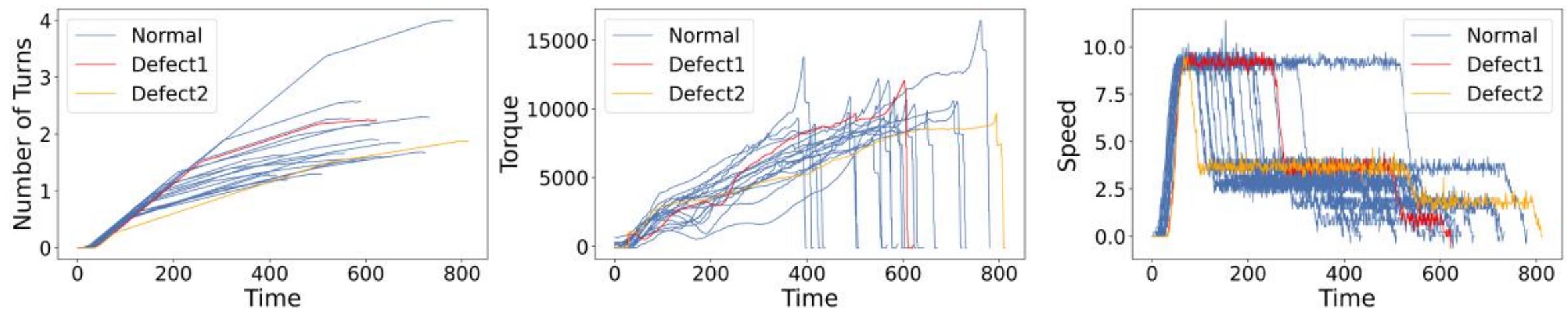

Figure 2. 20 samples from  a threaded pipe connection assembly line.

Besides the imbalanced classification problems, challenges are also arising from the unique functional data characteristics for threaded pipe connections. One challenge is that the sensing data are multichannel and existing representation learning-based learning methods lack explainable modeling and interpretations. Although traditional feature extraction methods for sequential data can automatically learn representations for classification, the learned representations are not interpretable for the multichannel functions. Another challenge is the incomplete observation of each process, which results in partially observed functional data for each sample. Since sensors are only installed to record threaded pipe connection process, the pre-connection process depicted in Figure 1 (a) is unobserved. Additionally, the inconsistency of the pre-connection process results in varying-length observations for each threaded pipe connection process, as shown in Figure 2. Thus, the proposed methodology should accurately classify each sample with consideration of partially observed functional data.

The practical challenges described above are common in production lines, and the defect classification problem becomes even more challenging when the challenges are present simultaneously. Learning from multichannel partially observed functional data is a novel topic in the context of imbalanced classification. There have been successful applications of imbalanced classification methods for defect classification in real industries, such as chemical reaction processes (Jiang and Ge, 2020), semiconductor manufacturing (Park *et al.*, 2022b), and vehicle air intake systems (Peng *et al.*, 2022). However, the current imbalance classification methods cannot be directly applied for multichannel partially observed functional data modeling. Moreover, the imbalanced classification problem needs to take into account the influence of learning from incomplete samples due to partial observability of the process. In the area of functional data analysis, there has been a functional neural network (Yao *et al.*, 2021) to learn from multichannel functions with varying-length observations. Prior research has also addressed challenges associated with partially observed functional data (Delaigle and Hall, 2013; James and Hastie, 2001; Kneip and Liebl, 2020; Kraus, 2015). However, these methods assume that the domain range for each sample is known, and the pre-specification of domain ranges is required, which may not be practical for data from real production lines, such as the threaded pipe connection process in our case.

To overcome the challenges of classifying the imbalanced, multichannel, and partially observed functional data, we propose a novel framework combining a functional neural network with deep metric learning (COMPILED). Specifically, a functional neural network is designed to directly learn fixed-length representations from multichannel functional data of varying lengths. To consider unobserved manufacturing processes, we employ a functional basis to pad the functional data before network encoding. In addition, we introduce a contrastive loss function tailored for deep metric learning on highly imbalanced functional datasets, thus facilitating efficient network training.

The contributions of our COMPILED framework can be summarized as follows:

1. We propose a novel classification approach for highly imbalanced functional data from manufacturing. By employing contrastive learning to address the imbalanced multi-class classification challenge, our framework realizes the automatic detection of nonconforming pipe connections and defect identifications.
2. With development of novel functional neural networks based on convolution operation (FunctionalCNN), our framework can directly learn from the multichannel functional data. Our functional neural network surpasses traditional learning methods for functional data by enabling learnable and intricate transformations of functional data. The learned functional basis in the functional neural network is also explainable.
3. To handle the partially observed functional data, domain knowledge is incorporated into the padding mechanism of functional neural networks, thereby alleviating the influence of the unobserved parts of threaded pipe connection process. Moreover, the convolution-based functional basis can directly encode functional data without requiring pre-specification of domain ranges.

The remainder of the paper is organized as follows. In Section 2, we review related works in imbalanced classification and classification of multichannel partially observed functional data. Section 3 provides preliminaries of deep metric learning and functional neural networks. Section 4 presents a detailed illustration of our proposed framework, and Section 5 offers an analysis of the evaluation results obtained from a real dataset of threaded pipe connections. Finally, Section 6 concludes the paper.

# 2. Related Works

## 2.1 *Imbalanced Classification*

Considerable research has been dedicated to addressing the challenge of imbalanced classification. Traditional strategies for dealing with class imbalance are broadly categorized into two groups: class rebalancing and data augmentation. Class rebalancing methods aim to balance the influence of different labels on the decision boundary. One representative is cost-sensitive learning (Elkan, 2001), which assigns larger weights to the minority labels in the loss functions. Data augmentation methods generate synthetic samples for minority labels to alleviate the imbalance problem. Examples of data generation methods include sampling methods, such as random oversampling and synthetic minority oversampling technique (SMOTE) (Chawla *et al.*, 2002), and deep learning methods based on pre-trained generative adversarial networks (GAN) (Radford *et al.*, 2015). However, both class rebalancing and data augmentation methods lead to overfitting when defect samples are scarce.

In recent years, the representation learning methods have shown great success in imbalanced classification given limited training samples. One branch of representation learning methods first learns the unsupervised representations of normal samples, and the model is then used for defect samples (Mou *et al.*, 2023). Another example is untrained methodologies (Tao and Du, 2025, Tao *et al.*, 2023) to learn defects and normal profile together by using only one sample. However, unsupervised learning can be challenging in cases where samples have varying lengths and are partially observed, as in our manufacturing dataset.

Another promising branch is to learn the supervised representations of each label based on deep metric learning. A neural network encoder first maps input data into the feature space, and the contrastive loss is then used to train the encoder and guarantee that the features are discriminative. Although existing studies have proposed losses based on contrastive learning to address imbalanced datasets (Khosla *et al.*, 2020; Wang *et al.*, 2021; Zhu *et al.*, 2022), these losses are primarily designed for image or speech data, and limited research focuses on classification using functional data in the literature. Therefore, both the contrastive loss and the neural network encoder need to adapt to the functional data in our case.

Given the above categories of methods, we summarize the current literature as Figure 3 with our focus in red rectangle.

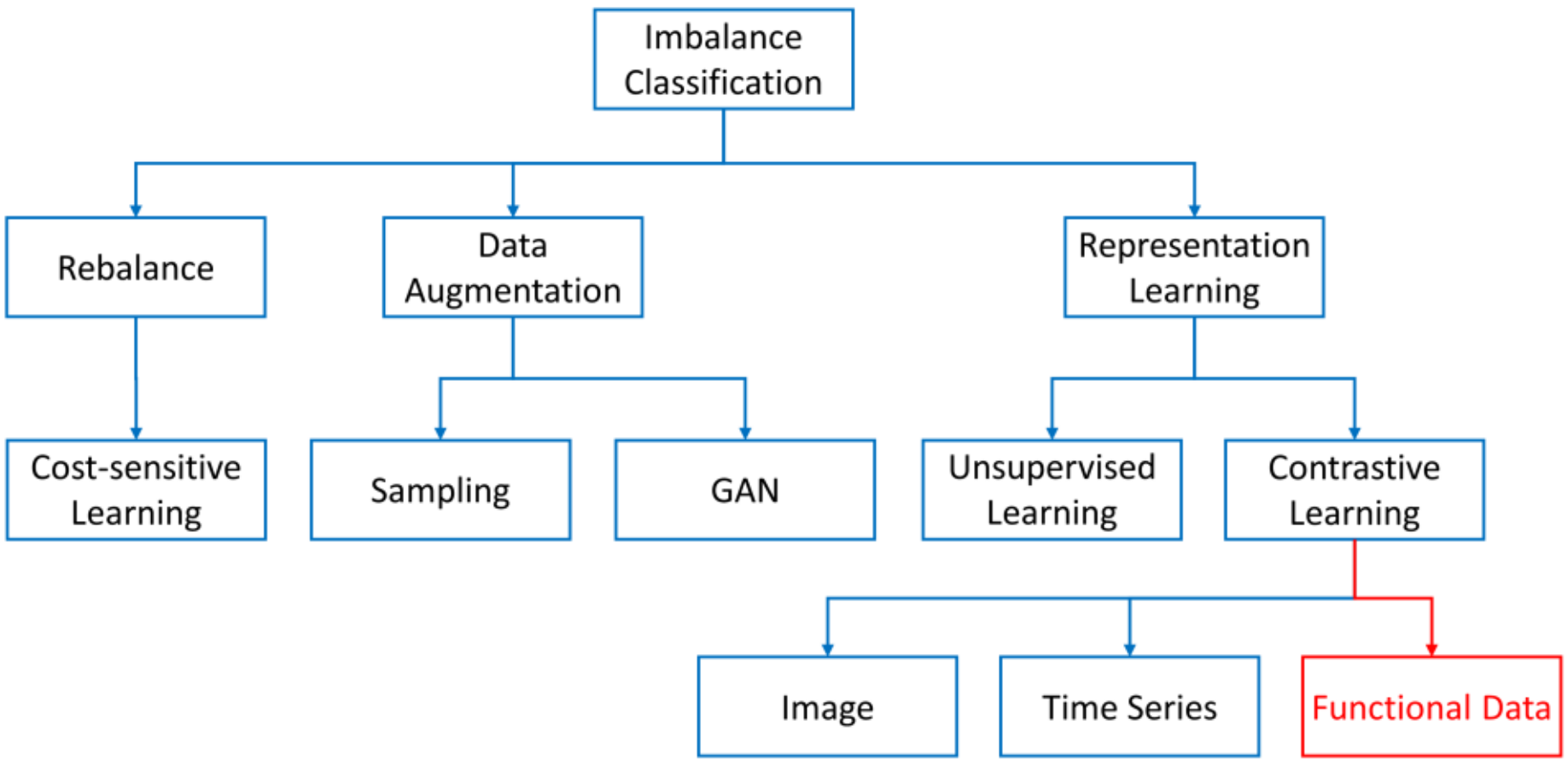

Figure 3. Methodology tree of imbalanced classification.

## 2.2 *Classification of Multichannel Partially Observed Functional Data*

Functional data is one type of high-dimensional data, where a process variable is characterized by a functional relationship with other process variables. Suppose that a dataset has $J$ labels and $N_j$ samples for label $j$. We use $\boldsymbol{x}(t), t \in [a, b]$ to denote one sample of functional data defined on a compact interval. Since only a limited number of observations can be observed from $\boldsymbol{x}(t)$, we use $\boldsymbol{X} \in \mathbb{R}^{T_{ij} \times C}$ to denote the $T_{ij}, i = 1, \ldots, N_j, j = 1, \ldots, J$ observations from multichannel function $\boldsymbol{x}(t)$ with $C$ channels.

Due to the production environment and sensor locations, the manufacturing process cannot be observed completely, such as our threaded pipe connection case. Therefore, the observations from the function $\boldsymbol{x}(t)$ are available only within a subset of $[a, b]$. Such datasets, known as partially observed functional data, are overlapped only within a specific observed interval $[a', b']$, where $a \leq a' < b' \leq b$.

To model and further classify the multichannel partially observed functional data, current methods can be classified into three major categories: distance measure-based, reconstruction-based, and representation learning-based methods.

Distance measure-based methods quantify the dissimilarity between samples by defining distance measures. The representative distance measures that can be applied to functional data include dynamic time warping (DTW) and integrated depth for partially observed functional data (Elías *et al.*, 2022). However, it is important to note that DTW (Müller, 2007) was originally designed for sequence alignment, and its ability to represent differences between different labels in a multi-class classification problem may be limited. Elías *et al.* (2022) assesses the centrality of functional data within a set of labeled functions. Nevertheless, Elías *et al.* (2022) is unsuitable for our case because it requires a pre-specified domain range for each sample. Furthermore, Elías *et al.* (2022) only treats multichannel functional data as a weighted sum of univariate functions, with the weights also requiring pre-specification.

Reconstruction-based methods aim to reconstruct the missing portions of a function and subsequently apply conventional classification techniques. Reconstruction is typically data-driven and based on fragments from other samples (Delaigle and Hall, 2013) or methods such as functional principal component analysis (FPCA) (Kneip and Liebl, 2020; Kraus, 2015). However, reconstruction is impractical when the domain range is unknown for each function. Moreover, data-driven reconstruction cannot leverage the physical mechanisms of the manufacturing process.

Representation learning-based methods directly transform functional data into low-dimensional representations with fixed lengths. Beginning with pioneering work (James and Hastie, 2001), various linear classifiers (Kraus and Stefanucci, 2019; Park *et al.*, 2022a) have been proposed. However, linear classifiers are not applicable because the assumptions of equal covariance for each label may not align with our dataset from the manufacturing process.

Apart from the limitations of each category of methods, most existing classification methods mentioned above focus primarily on binary classification, with limited exploration of imbalanced multi-class classification problems.

Recently, functional neural networks have shown great success in learning low-dimensional representations from functional data (Perdices *et al.*, 2021; Wang *et al.*, 2019; Yao *et al.*, 2021). With the power of neural networks, these approaches directly encode multichannel functional data with varying-length observations and perform more intricate transformation of functional data, leading to improved discrimination results. Another advantage of a functional neural network is that we can consider the label imbalance during training, which has not been explored by current classification methods in functional data analysis.

Since current classification methods for functional data cannot apply on imbalanced dataset and partially observed functional data without the pre-specified domains, we propose the neural network FunctionalCNN to deal with this kind of functional data and develop corresponding deep metric learning-based framework for the imbalance classification.

# 3. Preliminary

## 3.1 *Deep Metric Learning*

The process of imbalanced classification based on deep metric learning can be decoupled into two stages: supervised representation learning and classifier training. During supervised representation learning, an encoder maps input data into a discriminative feature space. In the subsequent training stage, the classifier is trained based on the learned representations.

When learning representations, contrastive loss is utilized to minimize the distance between data representations from the same labels while maximizing the distance between data from different labels in the feature space. Suppose we have one sample $i$, and its representation is denoted as $\boldsymbol{z}_i$. The representation from samples with the same label is denoted as $\boldsymbol{z}_p$, while the representation from the other samples is denoted as $\boldsymbol{z}_k$. The basic contrastive loss for sample $i$ is formulated as the following triplet contrast:

$$\mathcal{L}(i) = -log\frac{\exp(\boldsymbol{z}_i \cdot \boldsymbol{z}_p)}{\exp(\boldsymbol{z}_i \cdot \boldsymbol{z}_k)}. \tag{1}$$

In the settings of supervised learning, multiple samples are known to have the same label. Khosla *et al*. (2020) proposed the supervised contrastive loss when multiple positive samples and negative samples are present:

$$\mathcal{L}(i) = \frac{1}{|B_i| - 1}\sum_{p \in B_i \backslash \{i\}} -log\frac{\exp(\boldsymbol{z}_i \cdot \boldsymbol{z}_p)}{\sum_{k \in B \backslash \{i\}} \exp(\boldsymbol{z}_i \cdot \boldsymbol{z}_k)}, \tag{2}$$

where $B$ denotes the set of all training samples, $B_i$ denotes the set of training samples having the same label as sample $i$, and $B_i \backslash \{i\}$ is the set without element $i$. $|B_i|$ is the number of elements in the set $B_i$.

To deal with the imbalanced samples, Zhu *et al.* (2022) balanced the contribution of each label to the loss function by adding the weights for contrasting sample $i$ with negative samples:

$$\mathcal{L}(i) = \frac{1}{|B_i| - 1} \sum_{p \in B_i \setminus \{i\}} -log \frac{\exp(\boldsymbol{z}_i \cdot \boldsymbol{z}_p)}{\sum_{j=1}^{J} \frac{1}{|B_j|} \sum_{k \in B_j} \exp(\boldsymbol{z}_i \cdot \boldsymbol{z}_k)}, \tag{3}$$

where we suppose the training set $B$ has $J$ labels and the set of training samples under label $j$ is denoted as $B_j$. $|B_j|$ is the number of elements in the set $B_j$. Our contrastive loss is inspired by Equation (3) and part of the loss function is similar to Zhu *et al.* (2022).

## 3.2 *Functional Neural Network*

The purpose of the functional neural network is to use a set of functional bases to project the functional data $\boldsymbol{x}(t)$ and extract representations. Denote the $c$th basis function as $\boldsymbol{\beta}_{(c)}(t), c = 1, \dots, C_1$, the score of $\boldsymbol{x}(t)$ regarding to the basis $\boldsymbol{\beta}_c(t)$ is as follows:

$$\langle \boldsymbol{\beta}_c(t), \boldsymbol{x}(t) \rangle = \int \boldsymbol{\beta}_c(t)\, \boldsymbol{x}(t) dt, \tag{4}$$

where both $\boldsymbol{\beta}_c(t)$ and $\boldsymbol{x}(t)$ have infinite dimensions. Notably, we use the case when dealing with single-channel functional data as an illustration, and we only need to perform the same operation for each channel when dealing with multichannel functional data.

There are two kinds of functional neural networks that incorporate Equation (4) into neural networks. The first way is to use FPCA or pre-specified basis functions to model $\boldsymbol{x}(t)$ and use $\langle \boldsymbol{\beta}_c(t), \boldsymbol{x}(t) \rangle$ as the input of neural networks (Perdices *et al.*, 2021). The second way is to directly use neural networks to approximate the functional basis (Wang *et al.*, 2019; Yao *et al.*, 2021). The data encoded by the neural network is the observations of $\boldsymbol{x}(t)$, denoted as $\boldsymbol{X} = [\boldsymbol{x}(t_1), \dots, \boldsymbol{x}\left(t_{T_{ij}}\right)]^T$. The discrete version of Equation (4) is then as follows:

$$\langle \boldsymbol{\beta}_c(t), \boldsymbol{x}(t) \rangle = \boldsymbol{X}^T \boldsymbol{\beta}_c, \tag{5}$$

where the basis vector $\boldsymbol{\beta}_c \in \mathbb{R}^{T_{ij} \times 1}$ is expected to have equal length as $\boldsymbol{X}$. The basis function is learned through a multilayer perceptron (MLP), denoted as $\boldsymbol{M}(t), t \in [a, b]$. Given $\boldsymbol{X}$ with any length $T_{ij}$, the basis vector is adaptively derived from $\boldsymbol{M}(t)$ as $\boldsymbol{\beta}_c = [\boldsymbol{M}(t_1), \dots, \boldsymbol{M}\left(t_{T_{ij}}\right)]^T$. Compared to the first way to use basis as input of functional neural network, the selection of functional bases or a prior is not required, and the information contained in labels can be used during learning for the second way.

One limitation of current functional neural networks is that the MLP-based functional basis lacks shift-invariance to the input data and thus necessitates pre-specification of the domain range $[a, b]$ for each function. Although there has been efforts to develop functional neural networks with shift-invariant property to the input data (Heinrichs *et al.*, 2023), the domain range for each function and the pre-specified basis functions are still required. It is common for the functional data acquired in manufacturing cases to have varying-length observations without known domain ranges. Therefore, current functional neural networks can be further improved for manufacturing cases, such as our threaded pipe connection process.

## 3.3 *Convolution Layer and Dilated Convolution*

As the basic module of convolutional neural networks, the convolution layers perform the convolution operation on the input data. One benefit of convolution layers is that the network can deal with sequential data of any length. We suppose that one convolution layer has a kernel $\boldsymbol{w}$ with size $k_w$. The output $\boldsymbol{O}$ of the convolution layer given input data $\boldsymbol{X}$ is as follows:

$$\boldsymbol{O}_t = \sum_{k=t}^{t+k_w-1} \boldsymbol{w}_{k-t} \boldsymbol{X}_k . \tag{6}$$

Given the input data $\boldsymbol{X}$ of any length $T_{ij}$, the convolution filter will slide one observation of $\boldsymbol{X}$ each time until all the observations are iterated, and the output sequence is expected to have length $T_{ij} - k_w + 1$. Therefore, the padding operation is warranted to extend the input data and keep the length of output data unchanged.

Notably, the kernel size $k_w$ is an important hyperparameter for setting the convolution layer. Given an input long sequence, convolution layers with small kernel size can only capture the local information. The convolution layers with large kernel sizes can capture more global information, while the local information may be neglected and more computation burden will be brought. A good practice to determine the kernel size is to employ varying kernel sizes with dilated convolution (Oord *et al.*, 2016). When conducting dilated convolution, a convolution filter is applied over the input data and skip input values with dilations. According to current practices (Franceschi *et al.*, 2019; Oord et al., 2016; Yue *et al.*, 2022), the dilation size $d$ is usually doubled for each layer. Similar to basic convolution layers, the output length of the dilated convolution layers is still the input length $T_{ij}$ with the padding operation. The difference is that the receptive field of dilated convolution layers can grow exponentially with $d$. Thus, the dilated convolution can balance the extraction of local information and global information when dealing with the input data of large length $T_{ij}$.

Notably, the kernel size $k_w$ is an important hyperparameter when setting the convolution layer. Given an input long sequence, convolution layers with small kernel size can only capture the local information. The convolution layers with large kernel sizes can capture more global information, while the local information may be not considered. A good practice to determine the kernel size is to employ varying kernel sizes with dilated convolution (Oord *et al.*, 2016). Assuming a dilation convolution layer with dilation size $d$ and that the dilation is doubled for each layer, a convolution filter is applied over the input data and skip input values with step $2^d$. Similar to basic convolution layers, the output length of the dilated convolution layers is still the input length $T_{ij}$ with the padding operation. The difference is that the receptive field of dilated convolution layers can grow exponentially with $d$, as illustrated in Figure 4. Thus, the dilated convolution can balance the extraction of local information and global information when dealing with the input data of large length $T_{ij}$.

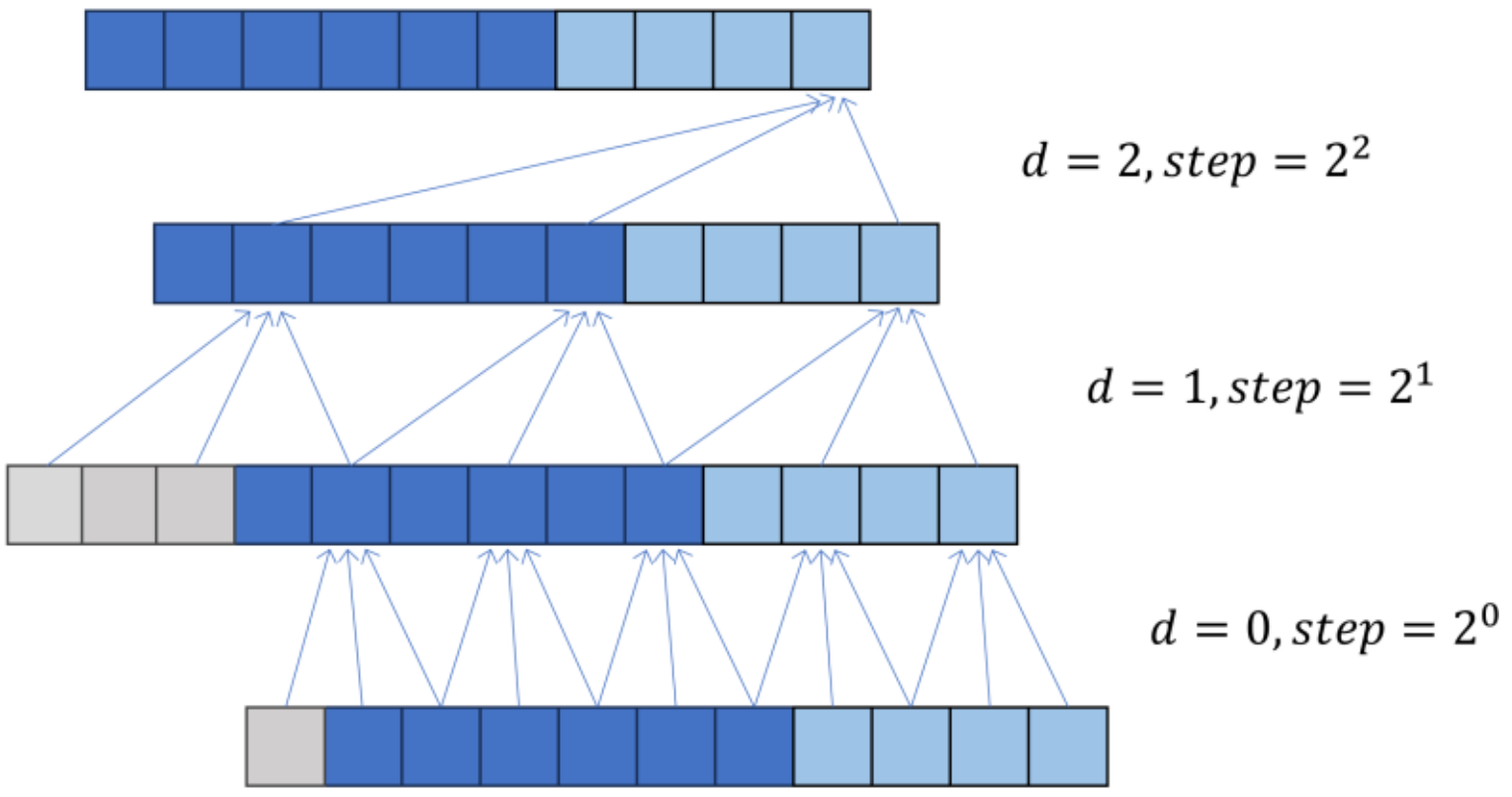

Figure 4. Illustration of dilated convolution.

## 4. Methodology

In our threaded pipe connection process, the multichannel and partially observed functional data necessitate the direct learning of dataset without relying on model assumptions or requiring pre-specification of domain ranges. Therefore, we propose the FunctionalCNN encoder capable of transforming functional data into low-dimensional features. Notably, the encoder can be trained to ensure that the resulting features are discriminative across different labels. Additionally, the challenge posed by the imbalanced nature of the dataset should be addressed effectively. Thus, we introduced a contrastive learning-based deep metric learning framework. The framework facilitates the training of the encoder using a highly imbalanced dataset, ultimately leading to a separable feature space where defect classification can be performed effectively.

Section 4.1 introduces the general framework. The functional neural network used to encode multichannel and partially observed functional data is elaborated in Section 4.2, and the contrastive loss designed for the imbalanced functional dataset is presented in Section 4.3. In Section 4.4, we provide guidelines for tuning the hyperparameters.

## 4.1 *General Framework*

This section introduces the proposed framework for the classification of imbalanced and partially observed functional data. The schematic diagram of the framework is shown in Figure 5. The general framework follows the paradigm of deep metric learning: each functional input $\boldsymbol{x}(t)$ is first encoded into a low-dimensional representation and then classified based on discriminative representation. As shown in Figure 6, $\boldsymbol{x}(t)$ first undergoes knowledge-infused padding, which reconstructs the padded area required for the neural network. The proposed functional neural network encoder then derives a fixed-dimensional representation. The initial representations of the samples from the different labels are mixed in the feature space. Therefore, a contrastive loss $\mathcal{L}_{contrast}$ is proposed, and the encoder is trained through backpropagation. After the representation learning stage, the learned representations under the same label are pulled together and the learned representations between different labels are pushed away. Finally, a classifier $\mathcal{L}_{classify}$, such as a support vector machine (SVM), is trained on the learned representations. The functional neural network encoder and the classifier $\mathcal{L}_{classify}$ after training are then used to predict the testing samples.

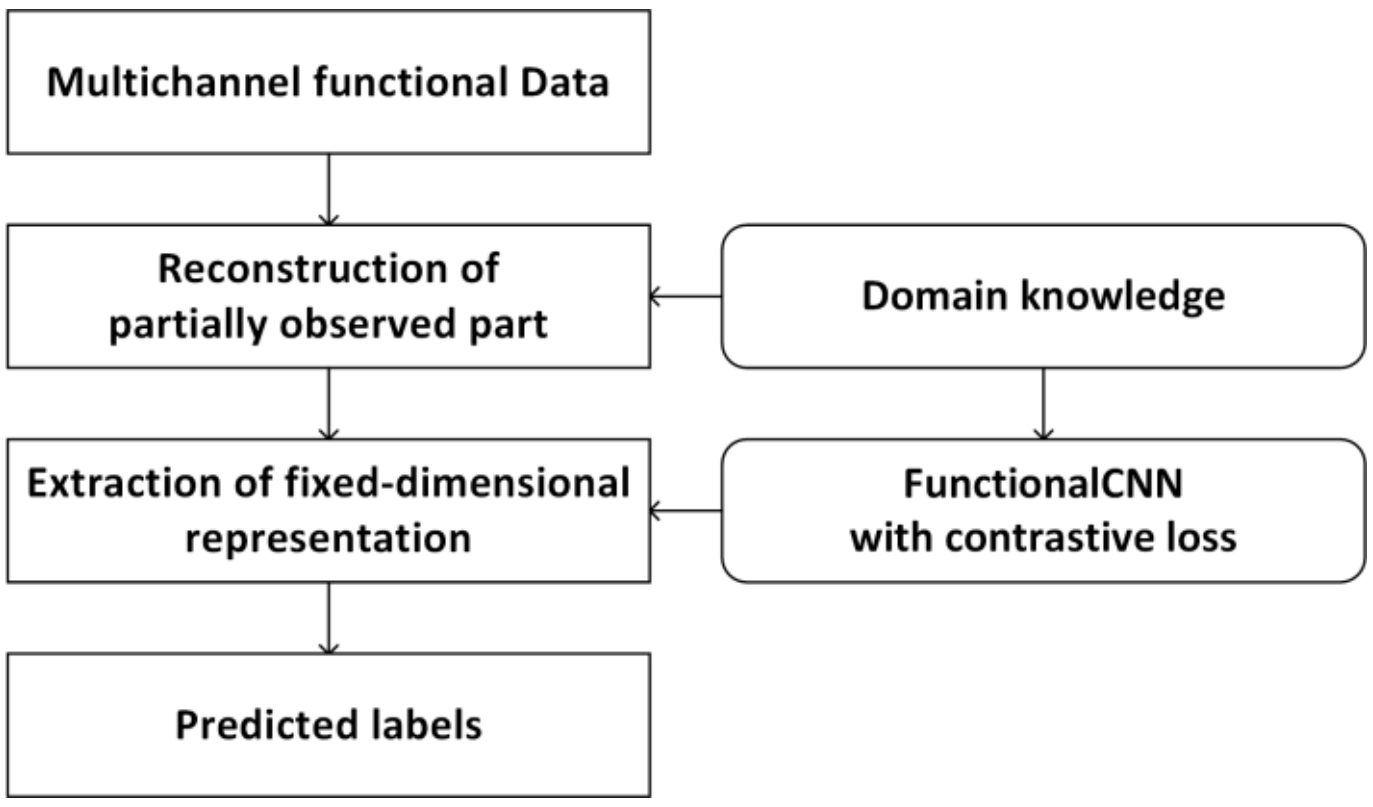

Figure 5. Schematic diagram of the framework.

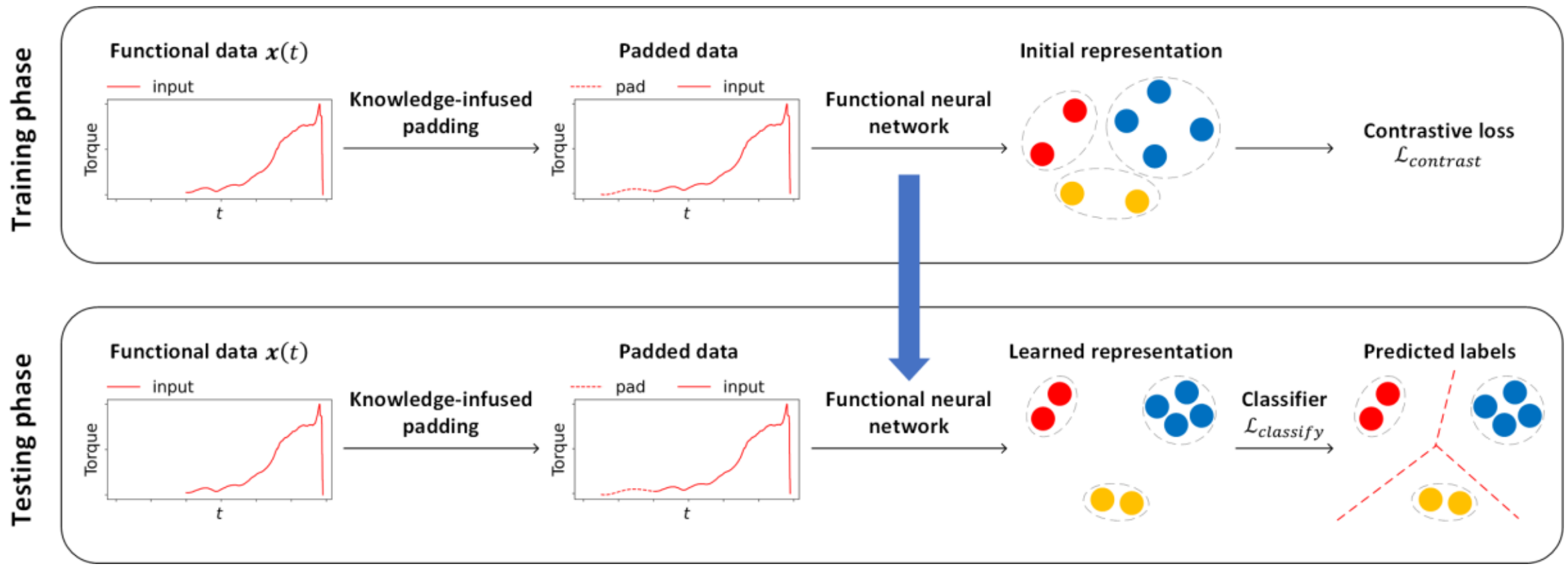

Figure 6. Training and testing phases of the framework.

The proposed functional neural network and contrastive loss can work collaboratively and enhance the performance of each other. The padding mechanism of a functional neural network considers the physical mechanism of the manufacturing process, which guarantees that the padded values do not impact the learned representation and subsequent contrastive loss. In return, contrastive loss enables the update of the functional neural network and improves the ability of the network to transform functional data after each iteration.

## 4.2 *Functional Neural Network Encoder*

As mentioned in Section 3.2, existing MLP-based functional neural networks (Wang *et al*., 2019; Yao *et al*., 2021) are not applicable in our case because the domain range for each function is unknown owing to partial observability. The neural network encoder should be shift-invariant and able to accept observations of functions with different ranges and lengths as input data. Therefore, we developed a functional neural network based on dilated convolution (Oord *et al.*, 2016) and knowledge-infused padding.

The benefits of dilated convolution include shift-invariant features guaranteed by the convolution mechanism and the ability to accept varying-length input data through dilated convolution. Although dilated convolution has shown promise for handling functional data of varying lengths, the padding mechanism remains a challenge. Traditional padding methods can introduce bias into the distribution of functions, which can negatively affect the performance of the neural networks. Another challenge is how to design a functional neural network with dilated convolution to deal with the unobserved parts of functional data. We illustrate the proposed solution for the two challenges in Section 4.2.1 and 4.2.2. The input data will first undergo the knowledge-infused padding is Section 4.2.1 and then the FunctionalCNN in Section 4.2.2.

### 4.2.1 *Knowledge-infused Padding*

In our case, the potential bias introduced by padding is a significant concern because the padding length increases exponentially with the dilation size in the dilated convolution. For example, the padding length is $(k_w - 1)2^d$ for dilation size $d$ when dilation is doubled for each layer. Therefore, the padded values will significantly influence the original distribution of the functional data. To address this challenge, we propose a novel approach called knowledge-infused padding. Similar to the reconstruction-based methods introduced in Section 2.2, our knowledge-infused padding attempts to reconstruct functions with a functional basis to generate the padded data. However, the reconstruction area in our framework is only the padded part in the convolution, rather than the entire missing part in terms of sensing data collection. The reason is that we assume the most discriminative patterns exist in the common domain, which is complete for every sample. After specifying the functional basis $\{\boldsymbol{\phi}_{(g)}(t)\}, g = 1, \ldots, G$, the knowledge-infused padding module is given in Algorithm 1 for any convolution layer with kernel size $k_w$ and dilation size. To distinguish the original functional data and the functional data after performing Algorithm 1, we denote the data as $\boldsymbol{x}_o(t)$ and $\boldsymbol{x}(t)$.

**Algorithm 1**. Knowledge-infused Padding

**Input**: Functional data $\boldsymbol{x}_o(t), t \in [a,b]$ with $T_{ij}$ observations; Convolution layer with kernel size $k_w$ and dilation size $d$; Average pooling layer with kernel size $k_w$; A set of functional bases $\{\boldsymbol{\phi}_g(t)\}, g = 1, \dots, G$;

**Output**: Padded data $\boldsymbol{x}(t), t = 1, \dots, T_{ij}{}'$

1: Obtain smoothed function $\overline{\boldsymbol{x}_o}(t) = \frac{1}{k_w}\sum_{s=t}^{t+k_w} \boldsymbol{x}_o(t), t = 1, \dots, T_{ij} - k_w$

2: Model $\overline{\boldsymbol{x}_o}(t)$ with functional basis $\overline{\boldsymbol{x}_o}(t) = \sum_{g=1}^{G} c_g \boldsymbol{\phi}_g(t)$

3: Estimate the weight $c_g, g = 1, \dots, G$ of each basis with FPCA

4: Interpolate $\overline{\boldsymbol{x}_o}(t)$ to length $T' = T_{ij} + 2(k_w - 1)2^d$

5: Obtain the padded data $\boldsymbol{x}(t)$:

6: Set $\boldsymbol{x}(t) = \overline{\boldsymbol{x}_o}(t), t \notin [a,b]$

7: Set $\boldsymbol{x}(t) = \boldsymbol{x}_o(t), t \in [a,b]$

Notably, the bases selection is a critical factor that directly affects the padding quality. Therefore, in this study, the selection of the functional basis is according to physical mechanisms of the manufacturing process. Specifically, if the manufacturing process exhibits a periodic pattern, the Fourier basis will be chosen to reconstruct the functional data collected from the process.

### 4.2.2 *Network Structure of FunctionalCNN*

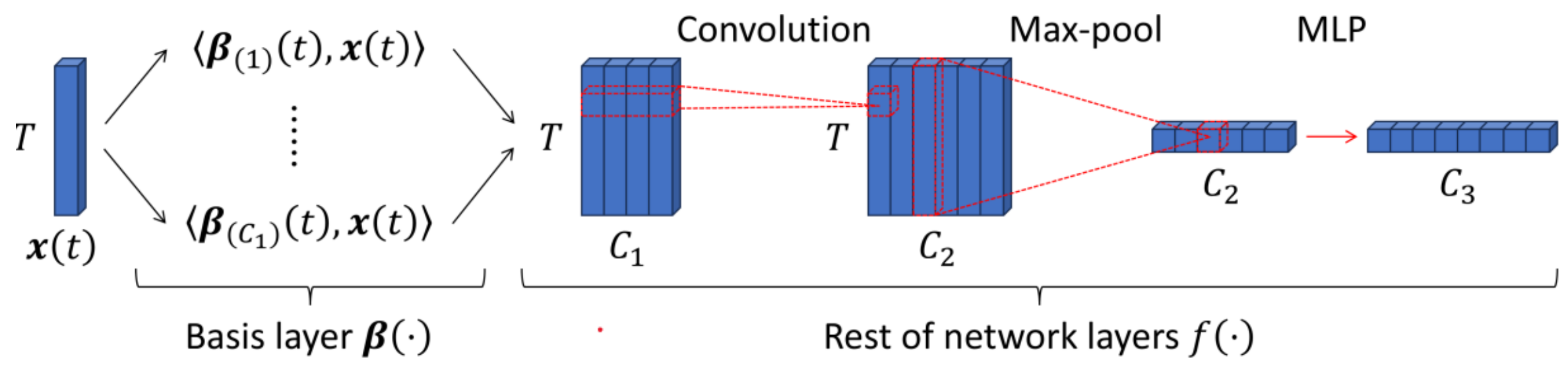

Figure 7. Structure of the functional neural network $f(\cdot)$.

To encode the functional data $\boldsymbol{x}(t)$ for downstream classification tasks, the encoder should consider the functional data structures while ensuring that the entire encoder is trained to extract the discriminative features. Our proposed FunctionalCNN achieves this by utilizing dilated convolution as the basic structure. The details of the network structure are shown in Figure 7.

Given the expected size $C_3$ of representations, the network shown in Figure 7 directly encodes the function $\boldsymbol{x}(t)$ to obtain the representation $\boldsymbol{z} \in \mathbb{R}^{1\times C_3}$ after knowledge-infused padding. Specifically, the network is composed of two parts: a linear transform with a basis layer $\boldsymbol{\beta}(\cdot)$ and a nonlinear transform with a traditional neural network $f(\cdot)$. The output for $\boldsymbol{x}(t)$ encoded by our functional neural network can be formulated as $f\left(\boldsymbol{\beta}(\boldsymbol{x}(t))\right)$.

---

**Algorithm 2**. Basis layer $\boldsymbol{\beta}(\cdot)$ based on dilated convolution

---

**Input**: Functional data $\boldsymbol{x}(t)$ and its observations $\boldsymbol{X} \in \mathbb{R}^{T\times 1}$; $D$ layers of dilated convolution $\left\{\boldsymbol{b}_c^{(l)}(t), \boldsymbol{h}_c^{(l)}(t)\right\}, c = 1, \dots, C_1, l = 0, \dots, D-1$;

**Output**: $\boldsymbol{\beta}(\boldsymbol{x}(t)) \in \mathbb{R}^{T_{ij}\times C_1}$

1. **for** $c = 1, \dots, C_1$:
2. Obtain $\boldsymbol{B}_c^{(0)} \in \mathbb{R}^{T_{ij}\times 1}$ and $\boldsymbol{H}_c^{(0)} \in \mathbb{R}^{T_{ij}\times T_{ij}}$ from convolution $\left\{\boldsymbol{b}_c^{(0)}(t), \boldsymbol{h}_c^{(0)}(t)\right\}$
3. Calculate $\boldsymbol{u}_c^{(0)} = \boldsymbol{B}_c^{(0)} + \boldsymbol{H}_c^{(0)}\boldsymbol{X}$
4. **for** $l = 1, \dots, D-1$:
5. Obtain $\boldsymbol{B}_c^{(l)} \in \mathbb{R}^{T_{ij}\times 1}$ and $\boldsymbol{H}_c^{(l)} \in \mathbb{R}^{T_{ij}\times T_{ij}}$ from convolution $\left\{\boldsymbol{b}_c^{(l)}(t), \boldsymbol{h}_c^{(l)}(t)\right\}$
6. Calculate $\boldsymbol{u}_c^{(l)} = \boldsymbol{B}_c^{(l)}(t) + \boldsymbol{H}_c^{(l)}\boldsymbol{u}_c^{(l-1)}$
7. **end for**
8. Set $\langle \boldsymbol{\beta}_c(t), \boldsymbol{x}(t)\rangle = \boldsymbol{u}_c^{(D-1)}$
9. **end for**
10. Set $\boldsymbol{\beta}(\boldsymbol{x}(t)) = \left[\langle \boldsymbol{\beta}_1(t), \boldsymbol{x}(t)\rangle, \dots, \langle \boldsymbol{\beta}_{C_1}(t), \boldsymbol{x}(t)\rangle\right]$

---

The basis layer $\boldsymbol{\beta}(\cdot)$ is implemented by stacking $C_1$ channels and $D$ layers of dilated convolution layers without activation. We denote the dilated convolution layers as $\left\{\boldsymbol{b}_c^{(l)}(t), \boldsymbol{h}_c^{(l)}(t)\right\}, c = 1, \dots, C_1, l = 0, \dots, D-1$ with bias $\boldsymbol{b}_c^{(l)}(t)$ and weight $\boldsymbol{h}_c^{(l)}(t)$. The procedures in $\boldsymbol{\beta}(\cdot)$ given the input single-channel functions are shown in Algorithm 2, and the network structure of $\boldsymbol{\beta}(\cdot)$ is given in Figure 8. When dealing with multichannel functions with $C$ channels, we need to perform Algorithm 2 on each channel of the multichannel function.

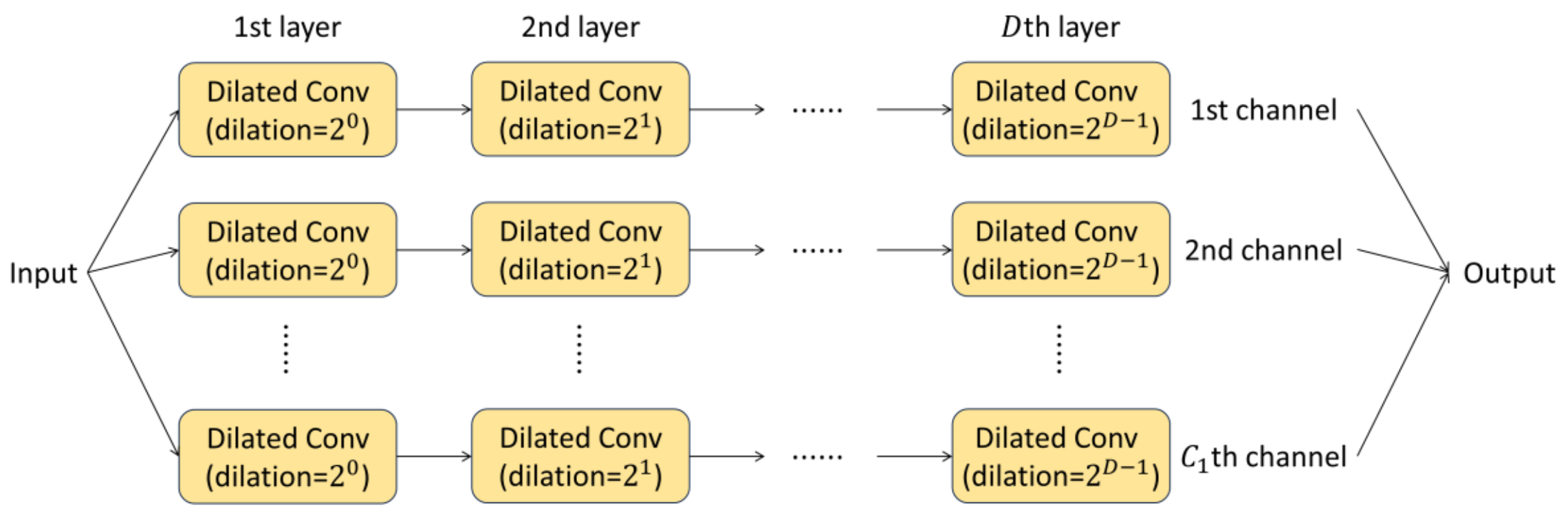

Figure 8. Structure of the basis layer $\boldsymbol{\beta}(\cdot)$.

The essence of the functional structure lies in the functional basis implemented by convolution operation. As shown in Algorithm 2, we do not add activation functions for the $D$ layers of the dilated convolution $\left\{\boldsymbol{b}_c^{(l)}(t), \boldsymbol{h}_c^{(l)}(t)\right\}$, which means that the entire basis layer $\boldsymbol{\beta}(\cdot)$ does not contain nonlinearity. By transforming the convolution weights $\boldsymbol{h}_c^{(l)}(t)$ into a Toeplitz matrix $\boldsymbol{H}_c^{(l)} \in \mathbb{R}^{T_{ij} \times T_{ij}}$ and bias $\boldsymbol{b}_c^{(l)}(t)$ as a vector $\boldsymbol{B}_c^{(l)} \in \mathbb{R}^{T_{ij} \times 1}$, we can derive $\langle \boldsymbol{\beta}_c(t), \boldsymbol{x}(t) \rangle = \boldsymbol{B}_c + \boldsymbol{H}_c \boldsymbol{X}$. $\boldsymbol{B}_c$ and $\boldsymbol{H}_c$ are derived as follows:

$$\boldsymbol{B}_c = \boldsymbol{B}_c^{(D-1)} + \boldsymbol{H}_c^{(D-1)} \boldsymbol{B}_c^{(D-2)} + \cdots + \left( \prod_{l=1}^{D-1} \boldsymbol{H}_c^{(l)} \right) \boldsymbol{B}_c^{(0)}, \quad (7)$$

$$\boldsymbol{H}_c = \prod_{l=0}^{D-1} \boldsymbol{H}_c^{(l)}, \quad (8)$$

where matrix $\boldsymbol{H}_c$ is the Toeplitz matrix for all $D$ convolution layers and can be viewed as a matrix basis. Therefore, the output of the basis layer $\boldsymbol{\beta}(\cdot)$ is still functional data. Compared with the MLP-based functional neural network (Wang et al., 2019; Yao *et al.*, 2021), $\boldsymbol{\beta}(\cdot)$ based on convolution makes the network shift-invariant to the input functions; Compared with current convolution-based functional neural network (Heinrichs *et al.*, 2023), $\boldsymbol{\beta}(\cdot)$ is adaptive and can accept the functional data without specifying the domain range and the basis functions.

Moreover, we can interpret the $C_1$ channels of convolution $\boldsymbol{\beta}_c(t)$ in $\boldsymbol{\beta}(\cdot)$ as $C_1$ functional bases. Similar to traditional functional modeling with a set of bases (Ramsay *et al.*, 2005), $C_1$ functional bases $\boldsymbol{\beta}_c(t)$ can be interpreted well through visualization and identification of the most important bases.

After transforming $\boldsymbol{x}(t)$ into $\boldsymbol{\beta}(\boldsymbol{x}(t))$ with basis layers, the neural network $f(\cdot)$ further transforms $\boldsymbol{\beta}(\boldsymbol{x}(t))$ into representation $\boldsymbol{z}$ through a convolution layer with $C_2$ channels and a $C_2 \times C_3$ MLP structure. The bias and weight for the convolution layer are denoted as $\boldsymbol{B}_f$ and $\boldsymbol{H}_f$, and the bias and weight for the MLP layer are denoted as $\boldsymbol{E}$ and $\boldsymbol{R}$. The procedure for nonlinear transformation $f(\cdot)$ is shown in Algorithm 3. The representation $\boldsymbol{z}$ is then used to calculate the loss functions in Section 4.3.

**Algorithm 3**. Nonlinear transformation $f(\cdot)$

**Input**: Output of basis layer $\boldsymbol{\beta}(\boldsymbol{x}(t)) \in \mathbb{R}^{T_{ij} \times C_1}$; Dilated convolution with bias $\boldsymbol{B}_f \in \mathbb{R}^{T_{ij} \times C_2}$, weights $\boldsymbol{H}_f \in \mathbb{R}^{C_1 \times C_2}$; MLP layer with bias $\boldsymbol{E} \in \mathbb{R}^{1 \times C_3}$ and weights $\boldsymbol{R} \in \mathbb{R}^{C_2 \times C_3}$;

**Output**: $\boldsymbol{z} \in \mathbb{R}^{1 \times C_3}$

1: Calculate $\boldsymbol{v}_1 \in \mathbb{R}^{T \times C_2} = \boldsymbol{B}_f + \boldsymbol{\beta}(\boldsymbol{x}(t))\boldsymbol{H}_f$

2: Calculate $\boldsymbol{v}_2 \in \mathbb{R}^{T \times C_2} = \text{LeakyReLU}(\boldsymbol{v}_1)$

3: Calculate $\boldsymbol{v}_3 \in \mathbb{R}^{1 \times C_2} = \text{AdaptiveMaxPool}(\boldsymbol{v}_2)$

4: Calculate $\boldsymbol{z} \in \mathbb{R}^{1 \times C_3} = \boldsymbol{E} + \boldsymbol{v}_3\boldsymbol{R}$

The details and proofs for constructing the Toeplitz matrix form of $\boldsymbol{B}_c$ and $\boldsymbol{H}_c$ in basis layer $\boldsymbol{\beta}(\cdot)$ are provided in Appendix A.1. We also illustrate how to visualize the functional bases learned from $\boldsymbol{\beta}(\cdot)$ in Appendix A.2. Guidelines for setting hyperparameters $D, C_1, C_2, C_3$ are also provided in Section 4.4.

## 4.3 *Contrastive Loss*

To train the functional neural network in Section 4.2, a loss function should be proposed for tackling label imbalance and data scarcity challenges when learning from the highly imbalanced functional dataset. We employ deep metric learning to train the neural network, and the contrastive loss $\mathcal{L}_{contrast}$ is used.

We first generate a mini-batch $B$ in each epoch through stratified sampling, and sample size $M_j$ for each label satisfies $M_1/N_1 = \cdots = M_j/N_j = \cdots = M_J/N_J$. Similar to Section 3.1, one sample $i$ is selected as the anchor sample and its representation is $\boldsymbol{z}_i = f\left(\boldsymbol{\beta}\left(\boldsymbol{x}^{(i)}(t)\right)\right)$. The representations of positive samples and negative samples are denoted as $\boldsymbol{z}_p = f\left(\boldsymbol{\beta}\left(\boldsymbol{x}^{(p)}(t)\right)\right)$ and $\boldsymbol{z}_k = f\left(\boldsymbol{\beta}\left(\boldsymbol{x}^{(k)}(t)\right)\right)$, respectively. $B_i$ denotes the set of samples having the same label as sample $i$ and $B_j$ denotes the set of samples under label $j$. $\mathcal{L}_{contrast}$ is then calculated after all samples in the mini-batch $B$ are encoded by $f(\cdot)$. The formula for $\mathcal{L}_{contrast}$ is given in Equation (9):

$$\mathcal{L}_{contrast}(i|Q) = \mathcal{L}_{inter}(i) + \mathcal{L}_{intra}(i|Q). \tag{9}$$

As shown in Equation (9), $\mathcal{L}_{contrast}$ is composed of an inter-class loss $\mathcal{L}_{inter}$ and an intra-class $\mathcal{L}_{intra}$ with smoothing kernels $Q$. Considering the imbalanced distribution of labels, $\mathcal{L}_{inter}$ in Equation (10) is the same as Zhu *et al.* (2022) which distinguishes the representations between imbalanced labels. For the contrast between the representations of anchor samples and negative samples $\exp(z_i \cdot z_k)$, the weights $1/|B_j|$ balance the influence of different sample size for each label.

$$\mathcal{L}_{inter}(i) = \frac{1}{|B_i| - 1} \sum_{p \in B_i \backslash \{i\}} -log \frac{exp(\boldsymbol{z}_i \cdot \boldsymbol{z}_p)}{\sum_{j=1}^{J} \frac{1}{|B_j|} \sum_{k \in B_j} exp(\boldsymbol{z}_i \cdot \boldsymbol{z}_k)}. \tag{10}$$

Regarding the data scarcity issue, we propose an intra-class loss $\mathcal{L}_{intra}$ in Equation (11) to avoid overfitting. To obtain more samples under one label, we augment the anchor sample $\boldsymbol{x}^{(i)}(t)$ and positive sample $\boldsymbol{x}^{(p)}(t)$, and the augmented functional data are denoted as $q(\boldsymbol{x}^{(i)}(t))$ and $q(\boldsymbol{x}^{(p)}(t))$. Notably, traditional augmentation methods used for images or time series may introduce bias into the distribution of functional data. We modify the augmentation method during the generation of the training samples to be suitable for functional data, aligning it with the characteristics of the manufacturing dataset. The augmentation is achieved by applying a set of smoothing kernels, denoted as set $Q$. As shown in Equation (12-13), each smoothing kernel $q \in Q$ is implemented through the average pooling layers with kernel size $k_q$, and we can set kernel set $Q$ in advance. Each contrast pair between the anchor sample and positive sample in $\mathcal{L}_{intra}$ are augmented with the same smoothing kernel.

$$\mathcal{L}_{intra}(i|Q) = \frac{1}{|B_i| - 1} \sum_{p \in B_i \backslash \{i\}} -log \frac{\frac{1}{|Q|} \sum_{q \in Q} exp\left(f\left(\boldsymbol{\beta}\left(q(\boldsymbol{x}^{(i)}(t))\right)\right) f\left(\boldsymbol{\beta}\left(q(\boldsymbol{x}^{(p)}(t))\right)\right)\right)}{exp(\boldsymbol{z}_i \cdot \boldsymbol{z}_k)}, \tag{11}$$

$$q(\boldsymbol{x}^{(i)}(t)) = \frac{1}{k_q} \sum_{s=t}^{t+k_q} \boldsymbol{x}^{(i)}(s), t = 1, \dots, T_{ij} - k_q, \tag{12}$$

$$q(\boldsymbol{x}^{(p)}(t)) = \frac{1}{k_q} \sum_{s=t}^{t+k_q} \boldsymbol{x}^{(p)}(s), t = 1, \dots, T_{ij} - k_q. \tag{13}$$

## 4.4 *Hyperparameter Tuning*

In this section, we present the guidelines for tuning the hyperparameters in our framework. The hyperparameters include $D, C_1, C_2, C_3$ in the functional neural network $f(\cdot)$ and the kernel set $Q$ in the contrastive loss $\mathcal{L}_{contrast}$.

The number of convolution layers $D$ and the number of bases $C_1$ determine the performance of basis layer $\boldsymbol{\beta}(\cdot)$. The ability to fit complex functions increases with increasing $D,$ whereas the padded length $(k_w - 1)2^{D+1}$ grows exponentially with $D$. Therefore, $D$ is set to satisfy the requirement that the padded length should not exceed the raw data length $(k_w - 1)2^{D+1} < T_{ij}$ for all the functions. The number of bases $C_1$ can first be set as an initial value $JCS$, where $S$ is the number of estimated piecewise linear segments for one function. During training, additional bases among the $C_1$ bases can be pruned by adding a channel-wise attention module to the neural network $f(\cdot)$. Channels $C_2$ and $C_3$ are the parameters of the traditional neural network $g(\cdot)$. Since the network structure of $g(\cdot)$ is the same as in previous practices (Franceschi *et al.*, 2019; Yue *et al.*, 2022), $C_2$ and $C_3$ can be set using the same settings in Franceschi *et al.* (2019); Yue *et al.* (2022). The setting of smoothing kernels $q$ in the set $Q$ can follow the procedure of smoothing functional data with a roughness penalty introduced in Ramsay *et al.* (2005).

# 5. Case Studies

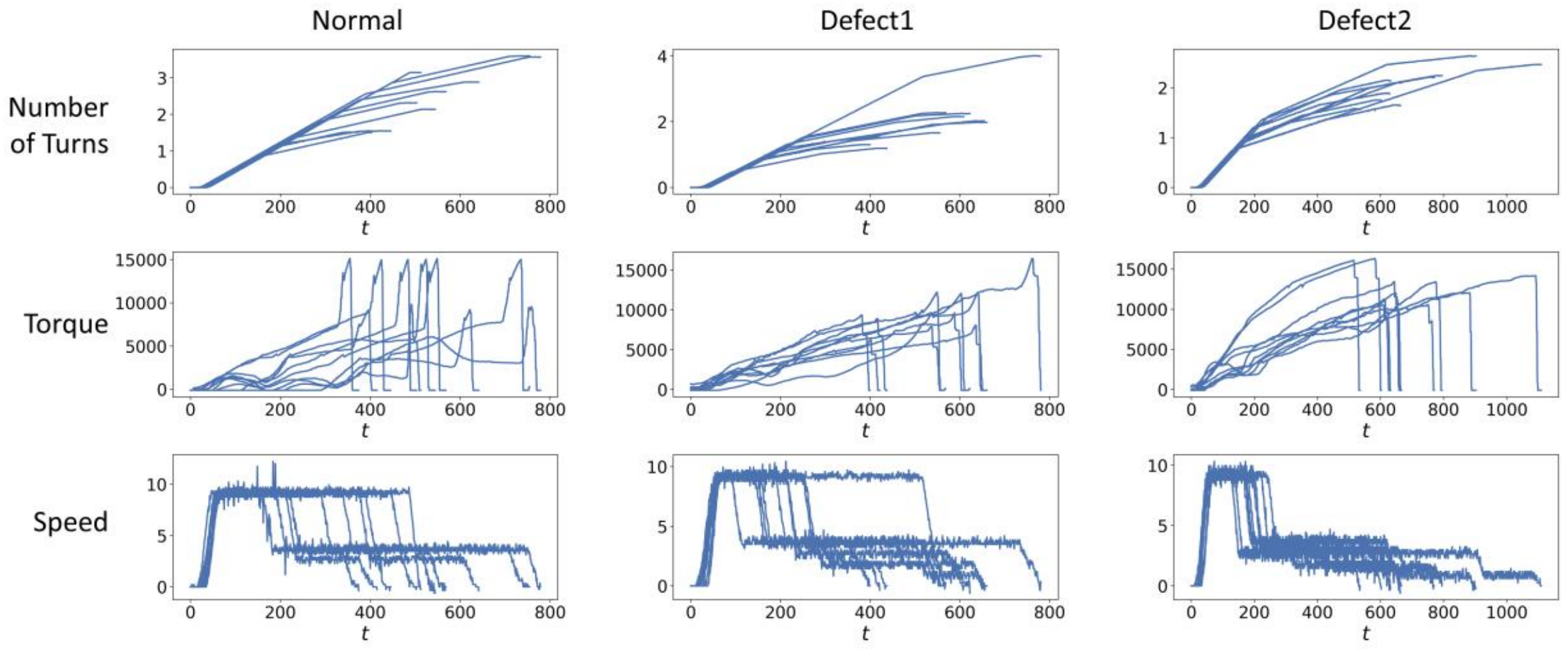

Figure 9. Multichannel Three channel functional data for each label in the steel pipe dataset under 3 labels.

In this section, we apply the proposed framework to a real dataset obtained from the threaded pipe connection process to demonstrate its effectiveness. A total of 658 samples are manually labeled by the manufacturer, including normal connections and two types of nonconforming connections. The dataset has a substantial class imbalance, with 599 samples for normal connections, 28 samples for one defect type, and 31 samples for another defect type. As shown in Figure 9, each sample is characterized by multichannel functional data, with varying-length observations during the thread engagement phase of the manufacturing process.

To configure our framework, we first select basis functions for the knowledge-infused padding introduced in Section 4.2.1. Taking the torque function in the manufacturing process as an example, we choose the Fourier basis because the missing observations of thread engagement process exhibits a periodic pattern. The results obtained using the Fourier basis and other traditional padding methods are shown in Figure 10. Compared with traditional padding methods, our knowledge-infused padding does not affect the distribution of the original functions. Similarly, we select the monomial basis for the number of turns and connection speed functions in the manufacturing process since both the number of turns and connection speed during the thread engagement process can be approximated with polynomials.

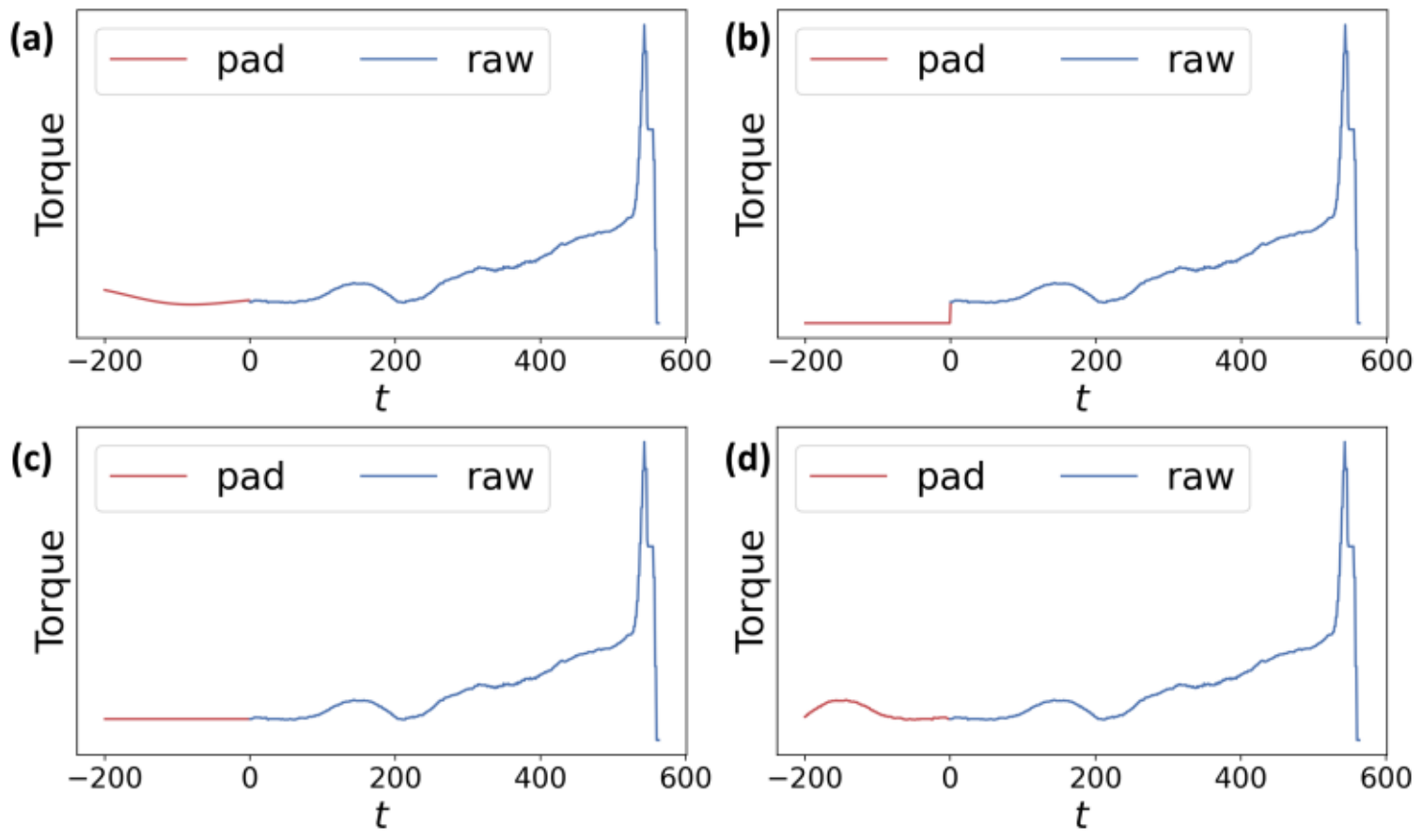

Figure 10. Padding results obtained using different padding methods. (a) Knowledge-infused padding; (b) Zero padding; (c) Replication padding; (d) Reflective padding

The other hyperparameters are set as follows: we set the number of convolution layers $D = 5$ and number of bases $C_1 = 40$, following the guidelines outlined in Section 4.4. The number of channels for other parts of the neural network, denoted as $C_2 = 160, C_3 = 320$, is consistent with the values used in Franceschi *et al.* (2019). Average pooling kernels of size $\{1,3,5\}$ are used in $Q$.

To demonstrate the efficacy of our framework when dealing with partially observed and imbalanced functional data, we visualize the representations of functional data, as shown in Figure 11. The fixed-dimensional representations are initially obtained from the multichannel and partially observed functions using functional neural networks, and are then visualized using the t-SNE technique (Van der Maaten and Hinton, 2008). We present the results from both the untrained network and the network after training to demonstrate the impact of representation learning. Figure 11 (a) illustrates that representations from different functional data samples are mixed in the feature space when using an untrained network. However, after representation learning through training with the contrastive loss in Section 4.3, the representations, as depicted in Figure 11 (b), become distinctly separated in the feature space. The separation facilitates subsequent classification, demonstrating that our framework enables the accurate classification of partially observed and imbalanced functional data.

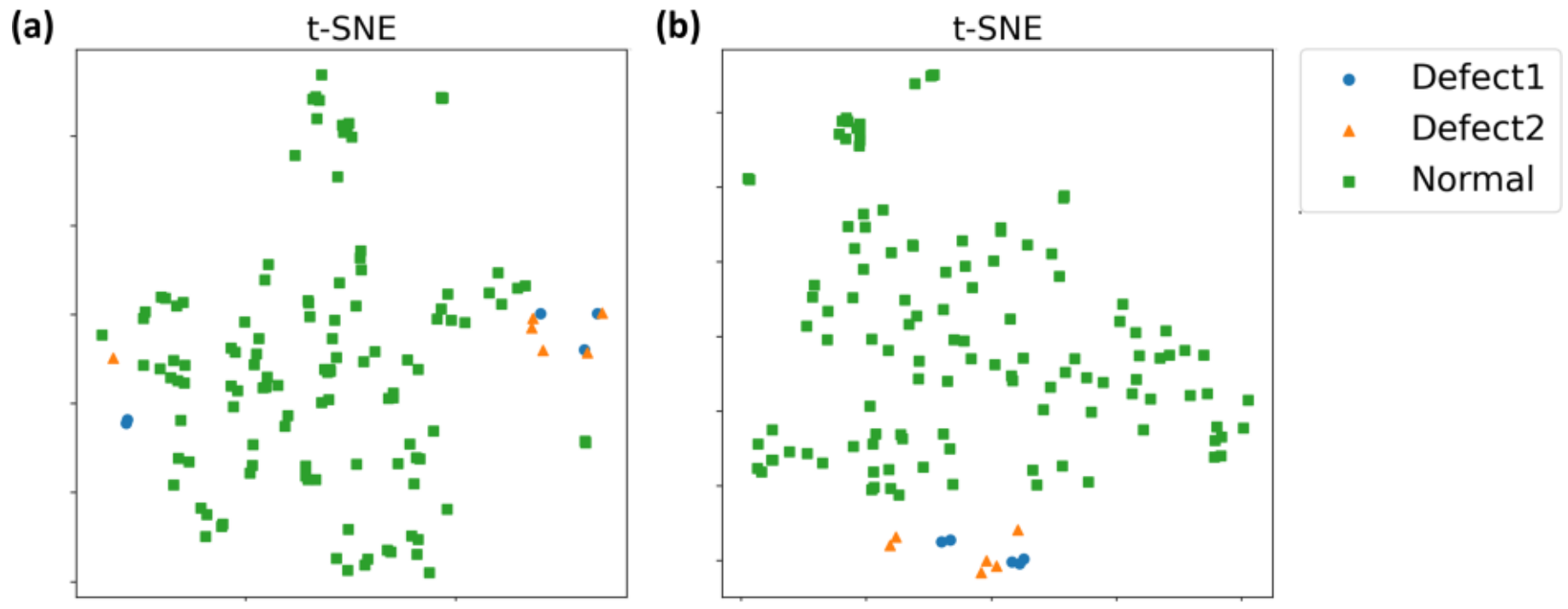

Figure 11. Visualization of the representations with t-SNE. (a) Representations from the untrained functional neural network. (b) Representations from the functional neural network after training.

For comparison, we select three benchmark methods, each representing one of the categories introduced in Section 2.2. The baseline method for classifying the functional data is DTW combined with a k-nearest neighbor (KNN) classifier. Additionally, we utilize Elías *et al.* (2022), a benchmark method for evaluating functional depth, which considers partial observability. The functional depth needs to be combined with depth-based classifiers (Cuesta-Albertos *et al.*, 2017; Li *et al.*, 2012) to classify functional data. To apply Elías *et al.* (2022) to our dataset, we use the R package proposed in Elías *et al.* (2022) to calculate depth measures and implement a depth-based classifier following Cuesta-Albertos *et al.* (2017). Sparse functional linear discriminant analysis (SFLDA) (Park *et al.*, 2022a) is chosen as a representative representation learning-based method, and its implementation is based on the open code available in Park *et al.* (2022a). To ensure fair comparisons, we adopt the same SVM classifier as the backbone for classifier training in both our framework and the depth-based classifier using Elías *et al.* (2022).

To evaluate the performance, we employ balanced accuracy and macro-F1 as evaluation metrics, which are commonly used for imbalanced classification. We perform a 5-fold cross-validation on the threaded pipe connection dataset. In each fold, we randomly select 25% of samples from the training set as the validation set. The average results of 5-fold cross-validation using our framework and competing methods are presented in Table 1. As shown in Table 1, our proposed framework outperforms the competing methods in terms of both balanced accuracy and macro-F1 metrics. To provide a more detailed analysis of the classification accuracy, we present the recall rates for each label obtained from 5-fold cross-validation in Table 2. Notably, all methods perform well in classifying normal samples. However, the distinguishing factor lies in the recall rates of defect labels. Our framework exhibits higher recall rates for minority labels than other methods, emphasizing its superiority in handling imbalanced datasets and effectively identifying defect types. The detailed classification results of each fold are shown in Appendix A.3.

Table 1. Comparison of performance on the threaded pipe connection dataset

| Framework | Balanced Accuracy | Macro-F1 |
|---|---|---|
| DTW | 0.750 | 0.744 |
| Elías *et al*. (2022) | 0.732 | 0.745 |
| SFLDA | 0.816 | 0.814 |
| COMPILED | **0.897** | **0.906** |

Table 2. Comparison of recall rates for each label

| Framework | Label | Fold 1 | Fold 2 | Fold 3 | Fold 4 | Fold 5 |
|---|---|---|---|---|---|---|
| DTW | Defect 1 | 0.50 | 0.50 | 0.83 | 0.80 | 0.60 |
| | Defect 2 | 0.50 | 0.83 | 0.67 | 0.71 | 0.33 |
| | Normal | 1.00 | 0.99 | 1.00 | 1.00 | 0.99 |
| Elías *et al*. (2022) | Defect 1 | 0.50 | 0.50 | 0.17 | 0.40 | 0.40 |
| | Defect 2 | 0.83 | 1.00 | 0.50 | 0.86 | 0.83 |
| | Normal | 0.99 | 1.00 | 1.00 | 1.00 | 1.00 |
| SFLDA | Defect 1 | 0.50 | 0.17 | 0.50 | 1.00 | 0.60 |
| | Defect 2 | 1.00 | 0.83 | 1.00 | 1.00 | 0.67 |
| | Normal | 1.00 | 0.98 | 1.00 | 1.00 | 1.00 |
| COMPILED | Defect 1 | 0.50 | 0.67 | 0.67 | 1.00 | 0.80 |
| | Defect 2 | 1.00 | 1.00 | 1.00 | 1.00 | 0.83 |
| | Normal | 1.00 | 1.00 | 1.00 | 1.00 | 1.00 |

## 6. Conclusion

This study introduces a novel deep metric learning framework tailored for the classification of imbalanced, multichannel, and partially observed functional data within the context of manufacturing processes. While prior research on classifying multichannel and partially observed functional data exists, addressing the challenge of imbalanced classification has not been explored within the realm of functional data analysis. To facilitate an imbalanced classification, we propose a novel functional neural network that encodes multichannel and partially observed functional data, allowing it to be trained effectively on highly imbalanced datasets. Furthermore, the padding mechanism and contrastive loss associated with the functional neural network are specifically tailored to address the characteristics of functional data derived from manufacturing processes. We also provide comprehensive guidelines for hyperparameter tuning.

To validate the effectiveness of our framework, we applied it to a real dataset from threaded pipe connection process. The results demonstrate that our framework outperforms existing benchmarks designed for partially observed functional data. This validation underscores the utility of our framework in achieving the accurate identification of nonconforming threaded pipe connections in industrial manufacturing processes.

## Acknowledgements

The authors acknowledge the generous support from Guangdong Basic and Applied Basic Research Foundation (No. 2023A1515011656), the National Natural Science Foundation of China (No.72001139, No.72371219, and No. 52372308), Guangzhou-HKUST(GZ) Joint Funding Program under Grant No. 2023A03J0651, and the Guangzhou Industrial Information and Intelligent Key Laboratory Project (No.2024A03J0628).